\documentclass{./archive_files/amia}
\usepackage{lipsum} 
\usepackage{amsmath}
\usepackage{amssymb}
\usepackage{mathtools}
\usepackage{amsthm}
\usepackage{hyperref}
\usepackage{xcolor}
\usepackage{xspace}
\usepackage{balance}
\usepackage{multicol}
\usepackage{multirow}
\usepackage{booktabs}
\usepackage{enumitem}
\usepackage{float}
\usepackage{listings}
\usepackage{inconsolata}  
\usepackage{longtable}
\usepackage{setspace}
\usepackage{subcaption}
\usepackage{mfirstuc}
\usepackage[utf8]{inputenc}
\usepackage[most,breakable]{tcolorbox}
\tcbset{
  promptbox/.style={
    colback=blue!2!white,
    colframe=blue!60!black,
    boxrule=0.4pt, arc=2pt,
    left=8pt, right=8pt, top=6pt, bottom=6pt,
    fonttitle=\bfseries, breakable, enhanced
  }
}
\usepackage{fvextra}
\tcbuselibrary{skins}
\newcommand{\header}[1]{\noindent \textit{\capitalisewords{#1}}\newline}
\newcommand{\subheader}[1]{\noindent \textit{\makefirstuc{#1}}}
\newcommand{\modelname}{ClinNoteAgents}

\begin{document}





\title{ClinNoteAgents: An LLM Multi-Agent System for Predicting and Interpreting Heart Failure 30-Day Readmission from Clinical Notes}

\author{
Rongjia Zhou$^{1}$, 
Chengzhuo Li$^{1}$,
Carl Yang, PhD$^{1}$\footnote[1]{Corresponding Authors: Carl Yang (j.carlyang@emory.edu), Jiaying Lu (jiaying.lu@emory.edu).}, 
Jiaying Lu, PhD$^{1}\footnotemark[1]$
}

\institutes{
$^{1}$Emory University, Atlanta, GA, USA
}

\maketitle

\section*{ABSTRACT}
\textit{Heart failure (HF) is one of the leading causes of rehospitalization among older adults in the United States. Although clinical notes contain rich, detailed patient information and make up a large portion of electronic health records (EHRs), they remain underutilized for HF readmission risk analysis. 
Traditional computational models for HF readmission often rely on expert-crafted rules, medical thesauri, and ontologies to interpret clinical notes, which are typically written under time pressure and may contain misspellings, abbreviations, and domain-specific jargon.
We present ClinNoteAgents, an LLM-based multi-agent framework that transforms free-text clinical notes into (1) structured representations of clinical and social risk factors for association analysis and (2) clinician-style abstractions for HF 30-day readmission prediction.
We evaluate ClinNoteAgents on 3,544 notes from 2,065 patients (readmission rate=35.16\%), demonstrating high extraction fidelity for clinical variables (conditional accuracy \(\geq\)90\% for multiple vitals), key risk factor identification, and preservation of predictive signal despite 60–90\% text reduction.
By reducing reliance on structured fields and minimizing manual annotation and model training, ClinNoteAgents provides a scalable and interpretable approach to note-based HF readmission risk modeling in data-limited healthcare systems.}

\section*{INTRODUCTION}

Heart failure (HF) remains a major global health challenge, affecting more than 55 million individuals worldwide, with nearly 80\% of cardiovascular deaths occurring in low- and middle-income nations~\cite{yan2021GBDHF}. Approximately 25\% of HF patients are readmitted within 30 days~\cite{khan2021HFReadmission}, imposing substantial clinical and financial burdens on health systems. Multiple studies have identified diverse contributors to readmission of HF, including HF exacerbation\cite{Eltelbany2019CausesHFReadmission}, comorbidities such as \textit{chronic obstructive pulmonary disease}, \textit{chronic kidney disease}~\cite{Jain2023EtiologiesPredictors30DayHFReadmission}, and socioeconomic factors\cite{Reddy2019ReadmissionsHF}.
Given the complex, multi-factorial nature of HF, 
computational risk modeling for HF readmission often requires comprehensive longitudinal patient data~\cite{umehara2020factors}. 
Electronic health record (EHR) data, which capture demographics, diagnoses, laboratory results, and medications, have therefore become a central data source for studying HF outcomes, including 30-day readmission~\cite{mahajan2017validated,yu2024machine}.
While containing valuable multimodal health information for HF patients, EHRs are often incomplete or unavailable in developing countries~\cite{murray2022GBD} due to financial, technological, and organizational barriers \cite{bostan2024OpenSourceEHR}.  This challenge is particularly evident in developing countries in Asia and Africa. Many hospitals in Bangladesh and Indonesia continue to rely on handwritten or locally stored digital notes due to financial constraints, limited IT capacity, and poor interoperability infrastructure \cite{Hossain2025EMR,Taher2025HRIS}. Similarly, healthcare facilities in Kenya, Uganda, and Ghana frequently experience unstable internet connectivity and shortages of trained health informatics personnel \cite{Alzghaibi2025EHRS,Mensah2024EHR}. In these settings, unstructured clinical notes often serve as the primary source of documented patient information. Despite widespread adoption of EHR in the U.S., around 80\% of clinical information remains embedded in free-text notes \cite{Kong2019ManagingUnstructured}. Therefore, clinical notes offer a pragmatic strategy for constructing predictive models when access to structured EHR data is limited.


Early HF readmission models relied primarily on structured EHR data such as demographics, socioeconomic status, medical history, and laboratory measurements \cite{Frizzell2017, Pishgar2022}. More recent work has incorporated unstructured clinical notes 
to capture richer contextual and temporal information \cite {Liu2019readmission}, with hybrid models combining structured variables and note-derived embeddings yielding further gains \cite{Golas2018}. This shift underscores the potential of text-based modeling for early detection of readmission risk. Growing evidence also highlights the importance of social determinants of health (SDOHs), defined by the World Health Organization as the conditions in which individuals live and work \cite{whoSDOH2025}. SDOHs have been repeatedly linked to HF outcomes, with employment status, housing stability, and social support identified as major contributors to readmission risk.\cite{guevara2024llmsdoh}. However, SDOHs are rarely structured in EHRs, limiting their use in predictive systems. Recent advances in natural language processing (NLP) and large language models (LLMs) have enabled the extraction of both clinical and social risk factors from unstructured notes, with machine learning and transformer models achieving state-of-the-art performance in HF readmission prediction using discharge notes \cite{alnomasy2025HFReadmission, huang2020ClinicalBERT}. Summarization-based preprocessing is shown to further enhance predictive signal quality \cite{boll2025DistillNote}, and recent studies show that LLMs can extract SDOH with near-clinician accuracy \cite{consoli2025SDoHGPT, gu2025SBDHReader}. Despite these advances, most existing approaches rely on predefined SDOH taxonomies or focus solely on social factors, often overlooking clinical predictors for readmission risk\cite{shao2025MiningSDOH}. These gaps highlight the need for a unified framework that jointly extracts and harmonizes social and clinical determinants from discharge notes to enable scalable readmission modeling.

To address these limitations, we present ClinNoteAgents, a LLM-based modular framework that transforms unstructured discharge notes into structured, interpretable representations of clinical and social risk factors for HF 30-day readmission. Our framework integrates (1) a structural extractor of clinical and social risk factors for statistical analysis of their relationship to readmission outcomes, and (2) a summarizer that produces qualitative or mixed-evidence summaries for predictive modeling.
By reducing reliance on structured EHR and minimizing manual annotation, ClinNoteAgents enables scalable HF risk analysis from discharge documentation in data-limited healthcare settings.

\section*{METHOD}

\subsection*{Study Design}
In this study, we leverage discharge notes, a rich clinical text source containing patients' demographics, chief complains, comorbidities, clinical measurements, social determinants of health and other critical risk factors, as the primary data source for building computational risk analysis models of HF 30-day readmission. This design highlights the importance of clinical notes as both a practical alternative to structured EHR data in resource-limited settings and an under-exploited information source in data-rich health systems.
To achieve HF 30-day readmission risk analysis, it involves (1) predicting the readmission risk; and (2) identifying the driving factors of that risk. We therefore formalize the scientific problem as two interdependent subtasks.

\subheader{Heart failure readmission risk prediction.}
Formally, let $\mathbf{X}_i=\{x_1,x_2,\dots,x_l\}$ denote the discharge note of length $l$ textual sequence for the \textit{i}-th hospital admission, and let $y_i\in \{0,1\}$ be the binary indicator of whether the patient was readmitted within 30 days $(y_i=1)$ or not $(y_i=0)$. The goal is to learn a predictive model $f:\mathbf{X}_i\rightarrow y_i$, that predicts whether a patient will be readmitted within 30 days based on the input discharge note.




\subheader{Heart failure readmission risk factors mining.} 
To systematically quantify the associations between risk factors $\mathbf{R}_i$ and 30-day HF readmission outcome $y_{i}$, the first step is to identify a structured set of risk factors from each discharge note and then conduct statistical analyses to evaluate their relationships with readmission risk.
Specifically, the discharge note-based risk factors extraction process can be defined as $\mathbf{R}_i=h(\mathbf{X}_i)$, where \(h(\cdot)\) denotes the risk factor extractor, and $\mathbf{R}_i=\{(r_j,v_j)\}_{j=1}^{k}$ denotes a set of risk factors with each factor type $r_j$ associated with a risk factor value $v_j$. Examples of extracted tuples $(r_j,v_j)$ include (body temperature, 97), (condition, renal cell cancer), (employment, retired).
After obtaining structured risk factor representations $\mathbf{R}_i$ for each discharge note, the second step is to conduct statistical analysis to quantify their associations with 30-day readmission. Depending on the data type and distribution of each risk factor, appropriate statistical tests will be applied (e.g., chi-square test for categorical variables and logistic regression for continuous variables) to obtain both p-values and effect sizes. 

\subsection*{Data Source} 
This study used the publicly available Medical Information Mart for Intensive Care III (MIMIC-III) database~\cite{johnson2016mimic}. Following established conventions in prior studies~\cite{shao2025MiningSDOH}, heart failure (HF) patients were identified using ICD-9 diagnosis codes (398.91, 402.01, 402.11, 402.91, 404.01, 404.03, 404.11, 404.13, 404.91, 404.93, and all codes beginning with 428). 
For each patient with multiple hospitalizations, we constructed readmission pairs by linking each index admission with its subsequent readmission. The discharge summary from the earlier admission served as the input note $\mathbf{X}_i$ for modeling, while the outcome label $y_i$ was determined by the time interval between discharge and the next admission ($y_i=1$ if the interval $\leq 30$ days; otherwise $y_i=0$).
Table~\ref{tab:cohort_summary} summarizesthe cohort used in this study.

\begin{table}[H]
\centering
\caption{Summary statistics of the heart failure cohort.}
\vspace{-6pt}
\resizebox{\linewidth}{!}{
\begin{tabular}{ccccccc}
\toprule
\textbf{\# Patients} & \textbf{ReAdm (\%)} & \textbf{\# Notes} &
\textbf{\# Discharge Notes} & \textbf{\% Female} &
\textbf{Median Age (Q1--Q3)} & \textbf{Median LoS (Q1--Q3)}\\
\midrule
2065 & 35.16\% & 3604 & 3544 & 46.39\% &
72.9 (62.6--81.3) & 7.0 (4.0--13.0)\\
\bottomrule
\end{tabular}
}
\label{tab:cohort_summary}
\end{table}

\subsection*{Proposed LLM-based Multi-agent System}
\begin{figure}[H]
    \centering    
    \includegraphics[width=0.98\linewidth]{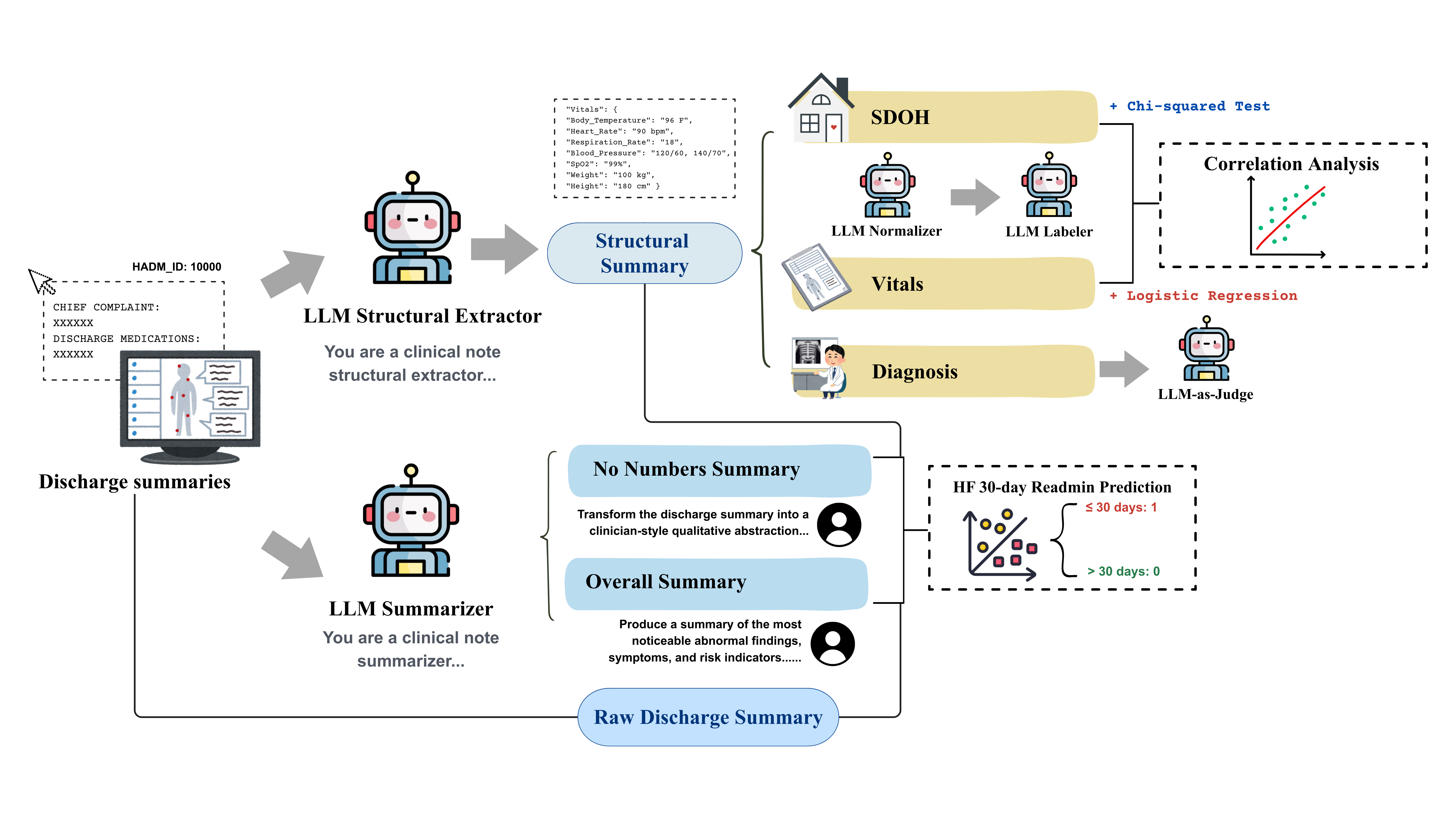}
    \vspace{-1cm}
    \caption{Overview of the ClinNoteAgents system for comprehensive clinical note analytics.}
    \label{fig:overview}
\end{figure}

We propose \modelname, an LLM-based multi-agent system for comprehensive clinical notes analytics. Our system comprises three agents: the Risk Factor Extractor, the Risk Factor Normalizer, and the Note Summarizer. The overall model framework is outlined in Figure~\ref{fig:overview}. 
All LLM agents were implemented using Qwen3-14B~\cite{yang2025qwen3} with thinking mode enabled. The extractor extracts clinical and SDOH variables; the normalizer and labeler standardize heterogeneous SDOH expressions into LLM-derived categories; and the summarizer produces clinician-style abstracts for downstream modeling.

\vspace{-8pt}

\subheader{Risk factor extractor.} We designed domain-specific LLM prompts to extract structured information from discharge notes across three categories: (1) charted SDOH—gender, age, primary language, and marital status; (2) uncharted SDOH—alcohol, tobacco, and drug use, transportation, housing, employment, parental status, and social support; and (3) clinical measurements—vital signs (temperature, heart rate, respiratory rate, oxygen saturation, height, weight, blood pressure), chief complaint, and diagnoses.

\begin{tcolorbox}[promptbox, title={Prompt example for LLM structural extractor}]
You are a clinical expert in extracting structured information from discharge summaries. Extract only the specific clinical factors that are important for predicting 30-day readmission in heart failure (HF) patients.

Output Schema Example:
\begin{Verbatim}[fontsize=\footnotesize, breaklines=true, obeytabs=true]
{
  "Charted_SDOHs": {"Gender": "M", "Age": "50", "Language": "null", "Marital_Status": "Married"},
  "NonCharted_SDOHs": {"Alcohol_use": "No","Tobacco_use": "1 ppd x 35y, quit 3 months ago","Drug_use": "No","Transportation": "null", "Housing": "null", "Parental": "null", "Employment_Status": "Retired", "Social_Support": "null"},
  "Clinical_Info": {
   "Vitals": {"Body_Temperature": "96 F", "Heart_Rate": "90 bpm", "Respiration_Rate": "18", "Blood_Pressure": "120/60", "SpO2": "99%", "Weight": "100 kg", "Height": "180 cm"}},
   "Chief_Complaint": {"Symptoms": "...", "Description": "..."},
   "Diagnoses": [{"Condition": "Renal cell cancer", "Details": "..."}, ...]
}
\end{Verbatim}
\end{tcolorbox}

\subheader{Risk factor normalizer.} Both charted and uncharted SDOH were often documented in heterogeneous free-text, limiting their use in quantitative analysis. To address this, we developed an LLM-based normalization module that standardized these entries into categorical values through a two-stage process: the first LLM (the normalizer) generated a concise set of normalized categories for each variable, and the second LLM (the labeler) assigned each extracted value to one of these categories. For charted SDOHs, only \textit{language} and \textit{marital status} required normalization, as \textit{gender} was already standardized (male and female). All uncharted SDOHs underwent the same normalization pipeline. Results were compared to existing reference taxonomies (MIMIC-SBDH~\cite{ahsan2021mimicsbdh} and LLM-SDOH~\cite{guevara2024llmsdoh}) using predefined categories.

\subheader{Note summarizer.} 
To assess whether LLM-based summarization improves HF readmission prediction, we implemented and compared two summarization methods: (1) an overall summary, which condenses the discharge note into a narrative emphasizing the most relevant clinical findings and risk indicators. Note that an overall summary can include numerical values, which contrasts to (2) a no-number summary, where all numerical values are replaced with qualitative descriptors to mitigate LLM instability in interpreting raw numeric data.

\begin{tcolorbox}[promptbox, title={Prompt example (overall summary)}]
You are a clinical expert in abstracting information from EHR discharge summaries. Given the discharge summary, produce a summary of the most noticeable abnormal findings, symptoms, and risk indicators that could be related to 30-day hospital readmission for heart failure (HF) patients.
\end{tcolorbox}


\begin{tcolorbox}[promptbox, title={Prompt example (no-number summary)}]
[\textbf{Same as overall summary above}]. Transform the discharge summary into a clinician-style qualitative abstraction, preserving section headers, while removing numbers and converting them to qualitative descriptors, focused on the potential 30-day readmission risk for HF patients.
\end{tcolorbox}

\section*{RESULTS}
\subsection*{Evaluation on HF Readmission Factors Mining}

We evaluated the risk-factor mining pipeline across extraction, normalization, and correlation analysis. Risk-factor mining transforms each discharge note into structured clinical and social variables and assesses their correlations with HF readmission. We first assessed extraction fidelity for clinical variables and charted SDOH using structured EHR as a surrogate ground truth, and applied an LLM-as-a-judge framework~\cite{huggingface2024llmjudge} for diagnoses. Next, we evaluated the LLM-based normalization module, which standardizes heterogeneous SDOH expressions into analyzable categories. Finally, we conducted statistical association analyses to identify risk factors significantly correlated with HF readmission.

\subheader{Structured EHR as surrogate ground truth.} 
Clinical measurements (vitals) and charted SDOH were evaluated against corresponding EHR tables to assess extraction coverage and conditional accuracy. 
Coverage, defined as the proportion of patients with non-null LLM-extracted values (``\% Extracted'' in Table~\ref{tab:ehr_agreement}), ranged from $4.03\%$ for \textit{Height} to $89.25\%$ for \textit{Heart Rate}. Among SDOHs, \textit{Gender} achieved the highest coverage ($99.18\%$), whereas \textit{Language} and \textit{Marital Status} were less frequently detected ($6.07\%$ and $25.73\%$, respectively). For each non-null extraction, values were compared with ground truth within variable-specific tolerance ranges (``Tolerance Range'' in Table~\ref{tab:ehr_agreement}) to compute conditional accuracy (``Cond Acc'' in Table~\ref{tab:ehr_agreement}). Tolerance ranges were designed to account for variability in clinical documentation, including rounding differences, unit conversions, or discrepancies between EHR and discharge notes. Mixed-unit variables, namely \textit{Temperature}, \textit{Height}, and \textit{Weight}, were compared in their native units when aligned. When discrepancies occurred, both LLM-extracted and structured EHR values were converted to canonical units ($^{\circ}\mathrm{C}$, cm, and kg) before comparison. Other vitals were consistently recorded and did not require unit harmonization.
For each vital within an admission, extracted and ground-truth values were converted to canonical units, aggregated, and summarized using medians. Mean absolute error and mean absolute percentage error (``MAE'' and ``MAPE'' in Table~\ref{tab:ehr_agreement}) were computed without tolerance adjustments to reflect raw deviation magnitudes.

\begin{table}[htbp!]
\centering
\caption{Agreement with structured EHR ground truth for extracted clinical variables and charted SDOH.}
\vspace{-6pt}
\begin{tabular}{lccccc}
\toprule
\textbf{Variable} & \textbf{\% Extracted} & \textbf{Tolerance Range} & \textbf{Cond Acc}& \textbf{MAE} & \textbf{MAPE} \\
\midrule
\multicolumn{6}{l}{\textit{Vitals}} \\
Temperature & 77.14\% & $\pm 0.5^{\circ}\mathrm{F}$, $\pm 0.3^{\circ}\mathrm{C}$ & 84.24\% & 1.40 & 1.42\% \\
HR          & 89.25\% & $\pm$ 5 bpm & 88.59\% & 11.42 & 13.74\% \\
RR          & 81.04\% & $\pm$ 1 breath/min & 93.77\% & 3.94 & 19.14\% \\
SPO2        & 85.92\% & $\pm$ 1\% & 94.16\% & 2.77 & 2.93\% \\
Height      & 4.03\% & $\pm$ 2 cm, $\pm$1 inch& 68.50\% & 2.91 & 1.73\% \\
Weight      & 15.74\% & $\pm$ 2 kg, $\pm$5 lbs& 57.90\% & 5.48 & 7.03\% \\
BP\_SYS     & 88.83\% & $\pm$ 5 mmHg & 90.85\% & 15.94 & 13.24\% \\
BP\_DIA     & 88.83\% & $\pm$ 5 mmHg & 91.24\% & 11.55 & 19.99\%\\
\midrule
\multicolumn{6}{l}{\textit{Charted SDOH}} \\
Gender          & 99.18\% &  --  & 99.94\% &  --  &  --  \\
Age             & 89.31\% &  --  & 93.38\% &  --  &  --  \\
Language        & 6.07\%  &  --  & 88.89\% &  --  &  --  \\
Marital\_Status & 25.73\% &  --  & 77.89\% &  --  &  --  \\
\bottomrule
\end{tabular}
\label{tab:ehr_agreement} 
\vspace{-6pt}
\end{table}

\subheader{Evaluation via LLM-as-a-judge.}
Diagnosis extraction was evaluated using an LLM-as-a-judge framework~\cite{huggingface2024llmjudge}, comparing LLM-extracted diagnoses with associated ICD codes. For each patient, the judge assigned a score from 0 (lowest) to 5 (highest) based on the semantic similarity between the extracted diagnoses and the ICD-9 codes. We computed two metrics: conditional accuracy, the proportion of correctly identified diagnoses among those extracted by the LLM, and absolute accuracy, the proportion of all ICD-9 diagnoses correctly recovered. As summarized in Table~\ref{tab:dx_extraction_metrics}, the LLM extracted fewer diagnoses per patient than the structured ICD-9 list. The average similarity score was 3.04, and the conditional accuracy was 62.27\%.

\begin{table}[htbp!]
\centering
\caption{LLM-as-a-judge evaluation of diagnosis extraction.}
\vspace{-6pt}
\begin{tabular}{cccccc}
\toprule
\textbf{Avg.\ \# ICD-9} &
\textbf{Avg.\ \# LLM-extracted} &
\textbf{Mean Score} &
\textbf{Median Score} &
\textbf{Cond Acc} &
\textbf{Abs Acc} \\
\midrule
15.14 & 5.91 & 3.04 & 3.00 & 62.67\% & 25.25\% \\
\bottomrule
\end{tabular}
\label{tab:dx_extraction_metrics}
\vspace{-6pt}
\end{table}

\subheader{Normalization results of clinical variables.}
LLM-based normalization agent was applied to reduce the high variability in free-text entries from SDOH extractions, thereby enabling direct correlation analyses. 
We first used k-medoids clustering to group semantically similar entries, selecting \textit{k = 200} to balance coverage and granularity (\textit{k = 300} produced overly fragmented clusters). Cluster medoids were then provided to an LLM to generate standardized category labels and descriptions, which a second LLM used to label each free-text entry as one of the normalized categories. The LLM-normalized categories of selected clinical variables are presented in Table ~\ref{tab:sdoh_categories}.

\begin{table}[htbp!]
\centering
\caption{LLM-normalized charted and uncharted SDOH categories.}
\vspace{-6pt}
\renewcommand{\arraystretch}{1.2}
\setlength{\tabcolsep}{5pt}
\begin{tabular}{p{3cm}p{12cm}}
\toprule
\textbf{Variable} & \textbf{Categories} \\
\midrule
Marital Status & Married; Widowed; Divorced/Separated; Single/Never Married; Unknown/Other \\
Alcohol Use & Abstinent/No Use; Current Heavy Use; Current Moderate/Social Use; Former Heavy Use; Former Moderate Use; Occasional/Rare Use; Unknown/Other; Past Use, Not Current \\
Tobacco Use & Never Smoker; Current Smoker; Former Smoker; Remote Tobacco Use; Occasional/Intermittent Use; Past Tobacco Use; High Pack-Year History; Unknown/Other \\
Transportation & Self-Driven; Non-Driver; Primary Transportation Method; Multiple Transportation Aids; Arranged Transportation; Assisted by Companion; Transportation Limitations; Unknown/Other \\
Housing & Living Alone; Living with Family Members; Institutional/Long-Term Care; Homelessness/Sheltered Living; Senior Housing/Retirement Communities; Residential Housing Type; Living with Non-Family Members; Home with 24/7 Care Services; Housing Instability/Unsafe Environment; Unknown/Other \\
Employment Status & Retired; Employed (Full-Time); Employed (Part-Time); Unemployed; On Disability; Self-Employed/Own Business; Student/Other Education; Unknown/Other \\
Social Support & Family Caregivers; Professional Caregivers; Social/Emotional Support; Living Arrangements; Lack of Social Support; Mixed Support Systems; Community/Non-Family Resources; Unknown/Other \\
\bottomrule
\end{tabular}
\label{tab:sdoh_categories}
\vspace{-1em}
\end{table}

\subheader{Correlation analysis.}
We evaluated associations across variable types, using logistic regression for numerical variables (vitals and age) and chi-square tests for categorical variables (charted and uncharted SDOH). Results for both analyses are reported in Table~\ref{tab:logistic_results} and Table~\ref{tab:chisq_results}.
Logistic regression identified three statistically significant variables: age, weight, and blood pressure. Age and BP are positively associated with HF readmission risk, whereas weight showed a negative association. The chi-square test showed housing as the only statistically significant SDOH variable.

\begin{table}[htbp!]
\vspace{0.5em}
\centering
\caption{Logistic regression correlation analysis for LLM-extracted clinical risk factors.}
\vspace{-6pt}
\begin{tabular}{lcccc}
\hline
\textbf{Variable} & \textbf{Coef} & \textbf{p-value} & \textbf{OR} & \textbf{OR 95\% CI} \\
\hline
Temperature & -0.019 & 0.650 & 0.981 & (0.904, 1.065) \\
HR           &  0.035 & 0.380 & 1.035 & (0.958, 1.118) \\
RR           &  0.047 & 0.239 & 1.048 & (0.969, 1.134) \\
SpO$_2$      &  0.006 & 0.871 & 1.006 & (0.933, 1.085) \\
Height       &  0.244 & 0.198 & 1.276 & (0.881, 1.848) \\
Weight       & -0.251 & \textbf{0.010} & 0.778 & (0.643, 0.942) \\
BP\_SYS      & -0.204 & \(<\)\textbf{0.001}& 0.816 & (0.738, 0.902) \\
BP\_DIA      & -0.085 & \textbf{0.037} & 0.918 & (0.847, 0.995) \\
Age& 0.001& \textbf{0.008} & 1.008&(1.002, 1.015)\\
\hline
\end{tabular}
\label{tab:logistic_results}
\vspace{-0.5em}
\end{table}

\begin{table}[htbp!]
\centering
\caption{Chi-square test results for LLM-extracted charted and uncharted SDOH.}
\vspace{-6pt}
\label{tab:chisq_results}
\begin{tabular}{lccccc}
\toprule
\textbf{Variable} & \textbf{N} & \textbf{Unique Values} & \textbf{Normalized Levels}& \textbf{Chi$^2$} & \textbf{p-value} \\
\midrule
\multicolumn{6}{l}{\textit{Charted SDOH}} \\
Gender          & 3515 & 2   & 2  & 2.62 & 0.106 \\
Language        & 214  & 41  & 5  & 3.50 & 0.478 \\
Marital Status  & 912  & 44  & 5  & 2.88 & 0.577 \\
\midrule
\multicolumn{6}{l}{\textit{Uncharted SDOH}} \\
Alcohol use     & 1930 & 851  & 8  & 6.94 & 0.435 \\
Tobacco use     & 2487 & 1517 & 8  & 10.28 & 0.173 \\
Drug use        & 864  & 315  & 5  & 3.56 & 0.468 \\
Transportation  & 37   & 36   & 8  & 9.08 & 0.247 \\
Housing         & 1031 & 554  & 10 & 21.13 & \textbf{0.012} \\
Parental        & 278  & 240  & 6  & 5.44 & 0.365 \\
Employment      & 1162 & 610  & 8  & 8.32 & 0.305 \\
Social support  & 954  & 769  & 8  & 6.90 & 0.439 \\
\bottomrule
\end{tabular}
\vspace{-1em}
\end{table}

\subsection*{Evaluation of LLM-based Note Summarization}
\begin{figure}[htbp!]
    \centering
    \includegraphics[width=0.7\linewidth]{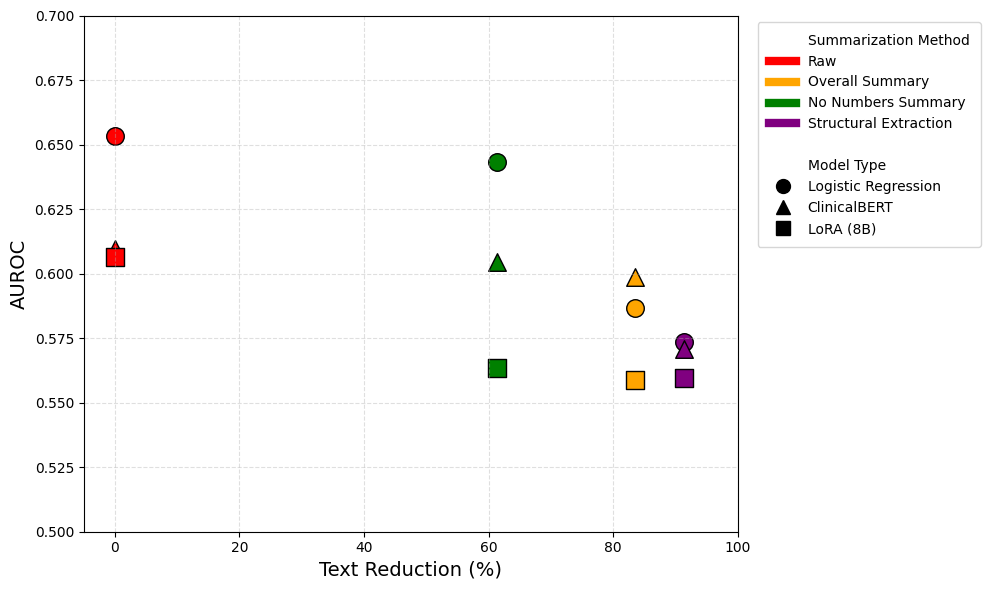}
    \vspace{-1em}
    \caption{HF 30-day readmission classification performance across summarization methods and model types.}
    \label{fig:tradeoff}
\end{figure}

We assessed whether transforming discharge notes into structured, clinician-style abstracts improves downstream readmission prediction. The summaries were generated using Qwen3-14B, and the prediction performance was evaluated using three classifiers: TF-LDF with logistic regression (LR), ClinicalBERT, and a LoRA-finetuned Qwen3-8B. The classification results are summarized in Figure~\ref{fig:tradeoff}.
Across all models, raw discharge notes achieved the highest AUROC (LR: 0.6535; ClinicalBERT: 0.6095; LoRA: 0.6064). Performance declines after summarization were moderate despite severe text reduction. Among the summarization methods, no-number summary (61.36\% word reduction) performed the best, with AUROCs remaining close to the raw baseline (LR: 0.6434; ClinicalBERT: 0.6046; LoRA: 0.5634). The overall summary (83.49\% word reduction) showed a larger decrease (LR: 0.5866; ClinicalBERT: 0.5986; LoRA: 0.5588) but remained competitive with ClinicalBERT. The structural extraction summary (91.44\% word reduction) produced the largest performance drop (LR: 0.5735; ClinicalBERT: 0.5708; LoRA: 0.5595).

\section*{DISCUSSION}

\subheader{LLM Risk Factor Extractor.} 
Accuracy within tolerance was high across most vitals. Conditional accuracy was highest for cardiorespiratory measures—\textit{SpO$_2$} (94.16\%), \textit{Respiratory Rate} (93.77\%), and \textit{Blood Pressure} ($\geq$90\%)—whereas anthropometric variables such as \textit{Height} (68.50\%) and \textit{Weight} (57.90\%) performed noticeably worse. These reductions were likely caused by heterogeneous reporting formats in discharge notes. Height may be recorded in centimeters or feet/inches, and weight in kilograms or pounds. 
Such variations may increase unit-conversion errors. Because LLMs exhibit unstable quantitative reasoning, and units in free-text form cannot be reliably validated, LLM-driven unit conversion was minimized during extraction. Nevertheless, values in uncommon units may still be misinterpreted. These patterns are consistent with evidence that LLMs hallucinate unsupported or incorrect clinical details during information extraction from discharge notes. \cite{das2025hallucination}. 
MAPE remained below 15\% for most vitals, indicating close agreement with the EHR surrogate ground truth. 
Small deviations occurred for \textit{Temperature} (1.40\%) and \textit{SpO$_2$} (2.77\%), while more variable measures such as \textit{Blood Pressure} (13.24\%-19.99\%) and \textit{Respiration Rate} (19.14 \%) showed larger discrepancies, consistent with their expected within-admission fluctuation. Several vitals and SDOHs showed low extraction coverage, largely due to under-documentation in discharge notes rather than extraction failure. Association analyses were therefore conducted only on patients with non-null values for each variable. While this approach may introduce selection effects, the findings should be interpreted as exploratory signals rather than population-level estimates. The LLM-as-a-judge evaluation showed that the model extracted fewer diagnoses per patient than the ICD-9 documentation. This is consistent with prior works that showed discrepancies between ICD coding and discharge notes \cite{kim2021icdaccuracy}. Despite the difference, the LLM achieved a moderately strong similarity score (mean 3.04) and a conditional accuracy of 62.67\%. This suggests that the extracted diagnoses were often semantically aligned with the ICD codes. These findings indicate LLM-based diagnosis extraction is feasible for scaling chart reviews, though discrepancies between clinical notes and coding systems warrant caution when using extracted diagnoses for downstream modeling.



\subheader{LLM Normalization and Correlation Analysis.}
The LLM-based normalization agent enabled correlation analysis that was otherwise infeasible on raw extractions. Prior works typically map SDOHs onto predefined taxonomies such as MIMIC-SBDH \cite{ahsan2021mimicsbdh} and LLM-SDoH \cite{guevara2024llmsdoh}. 
However, these schemes are coarse and often collapse clinically distinct expressions into a small set of categories (e.g., binary community factors and limited alcohol/tobacco categories). This limits cohort-specific nuance and may attenuate statistical associations. In contrast, the LLM-generated categories provided concise and cohort-specific representations while preserving relevant contextual detail. 
Logistic regression identified \textit{\textbf{age}}, \textit{\textbf{weight}}, and \textit{\textbf{blood pressure}} (BP) as significantly associated with HF readmission. Age showed a positive association, consistent with established epidemiology in which older patients are at higher risk of readmission \cite{shao2025MiningSDOH}. Systolic and diastolic BP exhibited negative associations, aligning with reports that low BP is associated with acute heart failure \cite{kim2024lowBP}. Lower discharge weight may similarly reflect HF risk. The observed correlations underscore the potential of LLM-based extraction to support early identification and monitoring of high-risk patients. Other vital signs were not significant, suggesting that their discharge-time values provide limited discriminative value for readmission. Chi-square analysis found housing to be the only significant SDOH. Other social factors commonly reported as influential, including smoking, social support, and marital status\cite{calvillo2013socialsupport}, were not significant in our cohort. These discrepancies reflect the under-documentation of SDOH in discharge notes or limitations in LLM extraction, highlighting the challenge of using discharge notes alone to capture SDOH.

\subheader{30-day Readmission Predication using LLM Summaries.}
LLM-generated summaries preserved most predictive signal needed for HF readmission modeling. Although summarization did not improve classification performance as initially expected, performance declines were modest across models, even with 60–90\% text reduction. This suggests the LLM preserved the majority of risk-relevant information, such as comorbidities, discharge stability, and key physiological measurements, while removing redundant content and noise. Substantial compression reduces token requirements for transformer models and produces more concise inputs for downstream review. Our results demonstrate that LLM summarization offers a balance between information preservation and computational efficiency, enabling scalable representation of clinical text while maintaining clinically meaningful signals for risk prediction.

\subheader{Model Choice in \modelname.}
We selected LLM agents over pre-trained language models (PLMs) such as ClinicalBERT because our note-to-structure tasks (extraction, normalization, and summarization) require generative reasoning and minimal supervision. LLMs are known to extract clinical entities and numeric data from unstructured clinical notes with high fidelity in zero- and few-shot settings \cite{agrawal2022LLMFewShot,adam2024ClinicalInfoExtraction}, enabling direct extraction of heterogeneous risk factors from discharge notes. LLMs also support SDOH extraction \cite{guevara2024llmsdoh} and, unlike PLMs, can perform normalization without requiring task-specific pretraining \cite{gu2025SBDHReader}. Moreover, LLMs provide substantially longer context windows than PLMs and is able to achieve summarization quality approaching expert benchmarks. \cite{vanveen2024AdaptedLLMsSummarization}.  We employed Qwen3 across agents to balance computing efficiency and performance, given its strong results in clinical information extraction benchmarks.  \cite{das2025hallucination}.

\subheader{Limitations.}
One limitation of our study is the indirect evaluation of LLM-extracted risk factors. For SDOH variables and most vital signs, we relied on structured EHR fields as surrogate ground truth and assessed extraction quality using tolerance-based accuracy and an LLM-as-judge approach. Another limitation is that we did not observe substantial performance gains from summarization alone. Finally, we did not include clinician-led evaluations of LLM extraction or summarization results, which limits formal assessment of clinical fidelity and real-world applicability.



\subheader{Ethical Considerations.}
This study involves secondary analysis of electronic health records and discharge summaries, all of which were fully de-identified in accordance with HIPAA standards. Access to the data required completion of the mandated training and certification, and all analyses were conducted under approved data-use conditions. Because de-identified clinical text may still contain sensitive contextual information, all large language models used in this work were deployed locally within secure computing environments. No patient data were sent to external, public, or third-party services.
Although LLM demonstrated reliable extraction performance, it remains prone to omissions and hallucinations. Inaccurate outputs could disrupt diagnostic reasoning, risk stratification, or care planning. Thus, LLM-based clinical systems should be used as decision-support tools, not standalone sources of truth. Responsible deployment requires rigorous validation and clear guardrails to ensure safe and appropriate use in clinical workflows.

\section*{CONCLUSION}
The study introduced ClinNoteAgents, an LLM–based multi-agent framework that operationalizes two core tasks: (1) HF readmission risk-factor mining and (2) HF readmission risk prediction. By transforming unstructured discharge notes into structured representations of clinical measurements, social determinants of health, and diagnoses, the system provides an end-to-end pipeline that directly supports predictive modeling and risk-factor analysis. For the readmission-factor mining task, the coordinated extraction and normalization agents recovered clinically relevant information with high fidelity, standardized heterogeneous medical term expressions, and enabled downstream statistical association analysis that would otherwise have been infeasible on raw free-text. These results demonstrate the feasibility of using LLMs to generate analyzable structural representations in limited structured EHR environments. For the readmission prediction task, abstractions produced by the LLM summarizer preserved most predictive signals from raw discharge notes despite substantial text compression. This indicates that the summarizer can distill long and noisy clinical narratives into concise representations that reduce computational cost while retaining clinical utility. Taken together, ClinNoteAgents provides a unified framework that enables scalable extraction, normalization, and summarization of free-text documentation to support HF readmission risk modeling. The system produces structured, interpretable abstractions compatible with rule-based pipelines, temporal reasoning modules, and clinical decision-support tools. In a clinical decision-support setting, ClinNoteAgents could be integrated into the EHR to generate readmission risk predictions alongside extracted risk factors and structured note summaries. Outputs would remain solely for advisory purposes and require clinician verification to mitigate hallucination and omission risks.

\section*{ACKNOWLEDGMENT}
This research was partially supported by the internal funds and GPU servers provided by School of Nursing's Center for Data Science of Emory University, Computer Science Department of Emory University, the US National Science Foundation under Award Numbers 2442172, 2312502, 2319449, and the US National Institute of Health under Award Numbers K25DK135913, RF1NS139325, R01DK143456 and U18DP006922.

\makeatletter
\renewcommand{\@biblabel}[1]{\hfill #1.}
\makeatother

\bibliographystyle{vancouver}
\bibliography{reference}

\end{document}